\documentclass[journal]{IEEEtran}
\usepackage{float,amsfonts,amsmath,graphicx,cite,bm,amssymb,amsthm,enumerate,epsfig,psfrag,cases,mathtools}
\renewcommand{\(}{\left(}
\renewcommand{\)}{\right)}
\renewcommand{\[}{\left[}
\renewcommand{\]}{\right]}

\renewcommand{\a}{\mathbf{a}}
\renewcommand{\b}{\mathbf{b}}

\renewcommand{\d}{\mathbf{d}}
\newcommand{\s}{\mathbf{s}}

\newcommand{\y}{\mathbf{y}}

\newcommand{\E}{\mathbf{E}}

\newcommand{\D}{\mathbf{D}}

\newcommand{\x}{\mathbf{x}}

\newcommand{\I}{\mathbf{I}}
\newcommand{\J}{\mathbf{J}}
\newcommand{\C}{\mathbf{C}}

\newcommand{\A}{\mathbf{A}}

\newcommand{\M}{\mathbf{M}}
\newcommand{\N}{\mathbb{N}}

\renewcommand{\u}{\mathbf{u}}
\renewcommand{\v}{\mathbf{v}}

\newcommand{\X}{\mathbf{X}}

\newcommand{\B}{\mathbf{B}}

\renewcommand{\b}{\mathbf{b}}

\newcommand{\Tr}[1]{{\rm{Tr}}\left(#1\right)}

\newcommand{\End}[1]{{\rm{End}}}

\renewcommand{\log}[1]{{\rm{log}}#1}
\renewcommand{\arg}[1]{{\rm{arg}}#1}

\renewcommand{\det}[1]{\left|#1\right|}

\renewcommand{\vec}[1]{{\rm{vec}}\(#1\)}

\newtheorem{theorem}{Theorem}
\newtheorem{prop}{Proposition}

\DeclareMathOperator*{\argmin}{\arg\!\min}
\newcommand{\norm}[1]{\left\lVert#1\right\rVert}

\usepackage{fontenc}
\usepackage{inputenc}
\usepackage[square,sort,comma,numbers]{natbib}
\usepackage[table]{xcolor}
\usepackage[ruled,vlined]{algorithm2e}
\usepackage{epstopdf}

\begin{document}
\title{Tyler's Covariance Matrix Estimator in Elliptical Models with Convex Structure}

\author{Ilya~Soloveychik,
        Ami~Wiesel

\thanks{Copyright (c) 2014 IEEE. Personal use of this material is permitted. However, permission to use this material for any other purposes must be obtained from the IEEE by sending a request to pubs-permissions@ieee.org}
\thanks{This work was partially supported by the Intel Collaboration Research Institute for Computational Intelligence, the Kaete Klausner Scholarship and ISF Grant 786/11.}
\thanks{The authors are with the Rachel and Selim Benin School of Computer Science and Engineering, the Hebrew University of Jeursalem, Israel (e-mail: ilya.soloveychik@mail.huji.ac.il)}
\thanks{The preliminary results were presented in ICASSP 2014.}
}

\maketitle

\begin{abstract}
We address structured covariance estimation in elliptical distributions by assuming that the covariance is a priori known to belong to a given convex set, e.g., the set of Toeplitz or banded matrices. We consider the General Method of Moments (GMM) optimization applied to robust Tyler's scatter M-estimator subject to these convex constraints. Unfortunately, GMM turns out to be non-convex due to the objective. Instead, we propose a new COCA estimator - a convex relaxation which can be efficiently solved. We prove that the relaxation is tight in the unconstrained case for a finite number of samples, and in the constrained case asymptotically. We then illustrate the advantages of COCA in synthetic simulations with structured compound Gaussian distributions. In these examples, COCA outperforms competing methods such as Tyler's estimator and its projection onto the structure set.
\end{abstract}


\IEEEpeerreviewmaketitle

\section{Introduction}
Covariance matrix estimation is a fundamental problem in the field of statistical signal
processing. Many algorithms for detection and inference rely on accurate covariance estimators \cite{krim1996two,dougherty2005research}. The problem is well understood in the Gaussian unstructured case. But becomes significantly harder when the underlying distribution is non-Gaussian, for example in elliptical distributions, and when there is prior knowledge on the structure. In this paper, we propose a unified framework for covariance estimation in elliptical distributions with general convex structure.

Over the last years there was a great interest in covariance estimation with known structure. The motivation to these works is that in many modern applications the dimension of the underlying distribution is large and there are not enough samples to estimate it precisely without additional structure hypotheses. The prior information on the structure reduces the number of degrees of freedom in the model and allows accurate estimation with a small number of samples. This is clearly true when the structure is exact, but also when it is approximate due to the well known bias-variance tradeoff. Prior knowledge on the structure can originate from the physics of the underlying phenomena or from similar datasets, e.g., \cite{fuhrmann1991application, roberts2000hidden, pollock2002circulant, stoica2011spice,wang1994adaptive}. %When the structure is defined using a closed convex set, a natural and computationally efficient solution is to project the naive unstructured estimators onto this set with respect to some predefined norm.

Many covariance structures are easily represented in convex form. Probably the most popular structure is the Toeplitz model. It arises naturally in the analysis of stationary time series which are used in a wide range of applications including radar imaging, target detection, speech recognition, and communication systems, \cite{snyder1989use, fuhrmann1991application, roberts2000hidden}. Toeplitz matrices are also used
to model the correlation of cyclostationary processes in periodic time series \cite{dahlhaus1989efficient}. In other settings the number of parameters can be reduced by assuming that the covariance matrix is sparse \cite{cai2012optimal,bickel2008regularized}. A popular sparse model is the banded covariance, which is associated with time-varying moving average models \cite{bickel2008regularized}. Another important example of a convex structure is the SPICE estimator, which was proposed in \cite{stoica2011spice} to treat high-dimensional array processing problems, where the covariance structure is approximated by a low-dimensional linear combination of known rank one matrices. In the last decade, all of these structures have been successfully considered in the Gaussian case.

In a different line of works, there is an increasing interest in robust covariance estimation for non-Gaussian distributions \cite{pascal2008covariance, abramovich2007time, wiesel2012geodesic, zhang2013multivariate, soloveychik2013group}. Significant attention is being paid to the family of elliptical and generalized elliptical (GE) distributions \cite{frahm2004generalized, frahm2007tyler}, which include as particular cases Generalized Gaussian distribution (GG), Compound Gaussian (CG) and many others \cite{frahm2004generalized}. Elliptical models are commonly used to measure radar clutter \cite{conte1991modelling}, noise and interference in indoor and outdoor mobile communication channels \cite{middleton1973man} and other applications. For these purposes robust covariance estimators were developed including Maronna's famous scatter M-estimator \cite{maronna1976robust}. Later Tyler \cite{tyler1987distribution} proposed a particular kind of scatter M-estimator which has become widely used \cite{abramovich2007time, pascal2008covariance,bandiera2010knowledge}. Although, generally M-estimators are given as solutions to optimization programs, Tyler showed that his M-estimator can be obtained as a solution to a simple fixed point equation. One of the most prominent disadvantage of these methods is that the optimization programs appearing from them are non-convex, thus making imposition of additional constraints rather difficult. In some cases it is possible to cure this obstacle by changing the metric of the underlaying manifold, thus appealing to the notion of geodesic ($g$-) convexity. It has been recently shown that some of the popular M-estimators are in fact $g$-convex, which significantly simplifies their treatment \cite{wiesel2012geodesic}. However, the imposition of additional constraints on the scatter matrix requires to define them as $g$-convex sets, rather than classical convex sets. Different improvements were achieved in this direction \cite{soloveychik2013group}, but this field is not still developed enough in order to solve the problems under consideration.

In the present work we derive COCA -  COnvexly ConstrAined Covariance Matching estimator. The underlying idea is based on the concept of Generalized Method of Moments (GMM) \cite{matyas1999generalized}. COCA searches for a covariance possessing the given convex structure that minimizes the norm of a sample moment's identity. This identity is in fact the optimality condition of an M-estimator. COCA tries to simultaneously satisfy this condition while constraining the structure. Unfortunately, this requires solving a high dimensional non-convex minimization program. Instead, we propose a convex relaxation and express COCA as a standard convex optimization with linear matrix inequalities which can be computed using off-the-shelf numerical solvers, such as CVX, \cite{cvx, gb08}. In terms of its analysis, we prove two promising results. First, in the unconstrained case, COCA is tight and identical to Tyler's estimator. This result basically ``convexifies'' Tyler's estimator. Second, in the structured case, COCA is asymptotically tight and hence consistent. Finally, we demonstrate the finite sample advantages of COCA over existing methods using numerical simulations.

The paper is organized in the following way. First, we formulate the problem, derive its Cramer-Rao performance bound and briefly describe the existing solutions: the sample covariance, Tyler's estimator and the projection method. We then introduce GMM, derive its convex relaxation named COCA and show that it coincides with Tyler's estimator in the unconstrained case. Then we prove that adding convex structure does not affect asymptotic consistency of the COCA estimator. Finally, we provide numerical examples and applications demonstrating the performance advantages of COCA.

We denote by $\mathcal{P}(p)$ the closed cone of hermitian positive semi-definite $p \times p$ matrices. We write $\M \succeq 0$ if $\M \in \mathcal{P}(p)$ for some $p \in \mathbb{N}$ and $\M \succ 0$ if in addition all the eigenvalues of $\M$ are positive. When convergence of random entities is considered $a.s.$ denotes the almost sure with respect to the probability measure convergence and $\xrightarrow{P}$ denotes convergence in probability. For a matrix $\M$, $\norm{\M}_F$ and $\norm{\M}_2$ denote correspondingly its Frobenius and spectral norms; $\norm{\M}$ stands for a not specified norm. $|\M|$ denotes the determinant of the matrix; $\M^T$ stands for the transpose and $\M^H$ for the conjugate transpose matrix. Given a matrix $\M$, the operator $\vec{\M}$ stacks all its columns into a one tall column. The sample measurements $\x_i \in \mathbb{C}^p$ are assumed to be independent and identically distributed (i.i.d.). We write $\x \sim \mathcal{N}(0,\M)$ for a centered complex circularly symmetric normally distributed random vector with covariance matrix $\M$. We use $j$ for the imaginary unit and avoid using it as an ordering index. $\I$ denotes the identity matrix of a proper dimension. Matrices are denoted by bold Capital letters $\M$, column vectors by bold non-capital $\v$ and scalars by non-capital $r$.

\section{Model and problem formulation}
\subsection{Complex Angular Elliptical distribution}
Consider a $p$ dimensional complex zero mean Generalized Elliptically (GE) distributed random vector $\s \in \mathbb{C}^p$, \cite{frahm2004generalized, frahm2007tyler}. Such a vector can be defined as \cite{frahm2004generalized}
\begin{equation*}
\s = r \bm\Lambda \u,
\end{equation*}
where $\u$ is a $q$ dimensional random vector, uniformly distributed over the unit complex hypersphere, $r$ is a nonnegative random variable, $\bm\Lambda \in \mathbb{C}^{p \times q}$. The random variable $r$ is called the generating variate of $\s$, we assume $r$ has no atom at $0$. If, in addition, we require $r$ to be stochastically independent of $\u$ the distribution becomes elliptical.
Below, we will normalize the random vectors and eliminate their generating variate. This will allow us to treat both GE and elliptical families in a similar way. The parameter $\bm\Theta_0 = \bm\Lambda {\bm\Lambda}^H$ is referred to as the dispersion or shape matrix of $\x$ and coincides with its covariance matrix (up to a scaling factor) when the latter exists. We assume $\bm\Theta_0 \succ 0$. The topic of this paper is the estimation of this shape matrix.

A closely related distribution is the Complex Angular Elliptically (CAE), \cite{greco2013cramer}, denoted by $\x \sim \mathcal{U}(\bm\Theta_0)$. This distribution can be obtained by normalizing a GE random vector $\s$:
\begin{equation*}
\x = \frac{\s}{\norm{\s}}, \: \s \neq 0,
\end{equation*}
The CAE probability density function is given by \cite{greco2013cramer}
\begin{equation}
p(\x) =   \frac{(p-1)!}{\pi^p} \frac{1}{|\bm\Theta_0|(\x^H  \bm\Theta_0^{-1}\x)^p},
\label{tyler_distr}
\end{equation}
where $\x$ belongs to a complex unit $p$-dimensional sphere. Ignoring additive constants, the negative log-likelihood function of CAE distribution is given by
\begin{equation}
l(\bm\Theta;\x) = \log|\bm\Theta| + p\log(\x^H  \bm\Theta^{-1} \x).
\label{neg_ll}
\end{equation}

The GE class of distributions includes proper complex Gaussian, compound Gaussian, elliptical, skew-elliptical, CAE and other distributions, \cite{frahm2007tyler}. An important property of the GE family is that the shape matrix of a population does not change whenever the random vector is divided by its Euclidean norm \cite{frahm2004generalized, frahm2007tyler}. As explained above, after normalization any GE vector becomes CAE distributed. This allows us to treat the shape matrices of all the mentioned distributions using a single robust estimator.

Note that the negative log-likelihood (\ref{neg_ll}) is insensitive to multiplication of the shape matrix by a positive constant, thus we are only interested in the estimation of the shape matrix up to a positive scalar factor. There are different approaches of fixing the scale; below we fix the trace of the estimator to get rid of this ambiguity.

\subsection{Structure}
In many applications, it is common to assume prior information on the structure of $\bm\Theta_0$. In particular, we can assume that it belongs to a known closed convex subset $\mathcal{S} \subset \mathcal{P}(p)$.
In many applications the role of $\mathcal{S}$ is played by a part of an affine hyperplane laying inside $\mathcal{P}(p)$. For simplicity, we consider the case of an affine set $\mathcal{S}$, but most of the results can be generalized to an arbitrary closed convex set separated from zero. Specifically, we assume that
\begin{equation}
\mathcal{S} = \mathcal{L} \cap \mathcal{P}(p),
\label{sset_def}
\end{equation}
where the hyperplane $\mathcal{L}$ is parametrized as
\begin{equation}
\mathcal{L} = \B_0 + \left\{ \sum_{i=1}^{k} a_i \B_i,\B_i \in \mathbb{C}^{p \times p}, a_i  \in \mathbb{R}\right\},
\label{affine_param}
\end{equation}
where $\B_0 \succeq 0$ and all $\B_i$ are hermitian, $k = \dim \mathcal{L} \leq p^2$. Note that $\mathcal{L}$ is parametrized by the real numbers $a_i  \in \mathbb{R}, i=1,\dots,k$, this is necessary to preserve the hermitian structure of the matrices in the affine hyperplane - if complex numbers are allowed the resulting sum may not be hermitian.

Typical examples of  such affine structures are:
\begin{itemize}
\item {\bf{Toeplitz}}: In stationary time series, the covariance between the $i$-th and the $h$-th components depend only on the the difference $|i-h|$. This kind of processes is encountered very often in many engineering areas including statistical signal processing, radar imaging, target detection, speech recognition, and communications systems, \cite{burg1982estimation,snyder1989use, fuhrmann1991application, roberts2000hidden, wiesel2013time,dahlhaus1989efficient}. The hyperplane $\mathcal{L}$ forms a $k=2p-1$ dimensional affine subspace. Using the notations in (\ref{affine_param}) we define the basis matrices as
\begin{equation*}
\B_0 = 0, \B_1 = \I,
\end{equation*}
\begin{equation*}
\B_i =  \begin{pmatrix}
  0 & \dots & 1 & \dots & \dots & 0 \\
  \vdots & \ddots & \vdots & 1 & \ddots & 0 \\
  1 & \dots & 0 & \dots & \dots & 0 \\
  0 & 1 & \dots & 0 & \dots & 1 \\
  \vdots & \ddots & \vdots & 0 & \ddots & \vdots \\
  0 & 0 & \dots & 1 & \dots & 0 \\
  \end{pmatrix}, i=2,\dots,p,
\end{equation*}
where the both $i-1$-th subdiagonals consist of ones, and analogously
\begin{equation*}
\B_i =  \begin{pmatrix}
  0 & \dots & j & \dots & \dots & 0 \\
  \vdots & \ddots & \vdots & j & \ddots & 0 \\
  -j & \dots & 0 & \dots & \dots & 0 \\
  0 & -j & \dots & 0 & \dots & j \\
  \vdots & \ddots & \vdots & 0 & \ddots & \vdots \\
  0 & 0 & \dots & -j & \dots & 0 \\
  \end{pmatrix},
\end{equation*}
$ i=p+1,\dots,2p-1.$

\item {\bf{Banded}}: A natural approach to covariance modeling is to quantify the statistical relation using the notion of independence or correlation, which corresponds to sparsity in the covariance matrix \cite{bickel2008regularized}. Assuming that $i$-th element of the random vector is uncorrelated with the $h$-th if $|i-h|>b$ leads to $b$-banded structure, also known as time varying moving average models. Using the definition in (\ref{affine_param}) we have symmetric matrices
\begin{equation}
\B_0 = 0, \B_{m} = \E_{ih} + \E_{hi},
\end{equation}
where $\E_{ih}$ are the unit matrices, $i$ runs from $1$ to $p$, $h$ from $i$ to $\min(i+b,p)$ and $m$ runs from $1$ to $m_s = \frac{(2p-b)(b+1)}{2}$ and insures linear ordering, and the hermitian ones
\begin{equation}
\B_{m} = j\E_{ih} - j\E_{hi},
\end{equation}
where $i$ runs from $1$ to $p-1$, $h$ from $i+1$ to $\min(i+b,p)$ and $m$ runs from $m_s+1$ to $2m_s - p = p(2b+1)-b(b+1)$.

\item {\bf{Direction of Arrival Problem}}: The problem of finding the direction of arrivals (DOA's) of $k$ plane waves impinging on a passive array of $p$ narrow-banded sensors can be reduced to that of estimating the parameters in the following model \cite{stoica2011spice}
\begin{equation}
\x_i = \B(\bm\theta)\y_i+\bm\omega_i,\quad i=1,\dots,n,
\end{equation}
where $\x_i \in \mathbb{C}^p$ are the noisy observation vectors, $\y_i \in \mathbb{C}^k$ are the unknown signal vectors, and $\bm\omega_i \in \mathbb{C}^p$ are hidden noise processes.  The vector $\bm\theta = [\theta_1\;\dots\;\theta_k]$ consists of unknown real parameters, and the matrix $\B(\theta) \in \mathbb{C}^{p\times k}$ has the following special structure:
\begin{equation}
\B(\bm\theta) = [\b(\theta_1) \dots \b(\theta_k)],
\end{equation}
where $\b(\theta_h)$ is the so-called steering vector or transfer vector (between the $h$-th signal source and the array output $i$). The exact form of the $\b(\theta)$ vectors depend on the array configuration. For example in a uniform and linear array we have
\begin{equation}
\b(\theta) = [1\; e^{j\theta}\; e^{2j\theta}\; \dots \; e^{(p-1)j\theta}]^T
\end{equation}
Assuming that the elements of $\y_i$ are statistically independent of the noise, the covariance matrix of $\x_i$ can be decomposed as
\begin{equation}\label{Tdoa}
\bm\Theta_0 = \sum_{i=1}^k  v_i^2 \b(\theta_i)\b(\theta_i)^H + \sigma^2 \I,
\end{equation}
where $v_i^2$ denote the signal sources powers and $\sigma^2$ stands for the power of the additive white noise. The goal is to estimate $\bm\theta$ from the measurements $\x_i,i=1,\dots,n$. In particular, a standard approach is to estimate the covariance with a structure that satisfies (\ref{Tdoa}) and solve for the corresponding $\bm\theta$. For this purpose, we generate a dense grid of $N$ points $\overline \theta_i$ over the interval of possible angles $[\theta_l;\theta_u]$ and fit the true covariance matrix by the linear model
\begin{equation*}
\bm\Theta = \sum_{i=1}^N  a_i \b(\overline\theta_i)\b(\overline\theta_i)^H + \sigma^2 \I,
\end{equation*}
where $a_i \geq 0$. If necessary, the $l_1$ norm of the parameter vector $\a = \{a_1,\dots,a_k\}$ can be constrained to ensure sparsity and linear independence. Returning to the structure notations in (\ref{affine_param}) we have
\begin{equation*}
\B_0=\sigma^2 \I, \B_i = \b(\overline\theta_i)\b(\overline\theta_i)^H,i=1,\dots,N.
\end{equation*}

%\item {\bf{Linear parameterization}}: Many other interesting models can be expressed as a linear combination of known matrices. In all such cases the framework developed in this work can be successfully applied.
\end{itemize}

\subsection{Problem}
We can now state the problem addressed in this paper: let $\x_i \sim \mathcal{U}(\bm\Theta_0), i=1,\dots,n, \; \bm\Theta_0 \in \mathcal{S}$ and assume the prior knowledge on the true covariance matrix is given in the form of an affine set. We are interested in estimation of the unknown shape matrix $ \bm\Theta_0$.

\section{Performance bounds }
Before addressing the possible solutions for the above covariance estimation problem, it is instructive to examine the inherent performance bounds. For this purpose, we consider the Cramer-Rao Bound ($\mathbf{CRB}$) on the variance of an unbiased estimators. Under mild regularity conditions, the $\mathbf{CRB}$ is asymptotically achievable by the Maximum Likelihood Estimator (MLE) and is therefore an important benchmark.

A straight forward approach to the $\mathbf{CRB}$ is to use the structured parameterization in (\ref{affine_param}) and compute the Fisher Information Matrix $\mathbf{FIM}(\a)$ associated with the parameter vector $\a = \{a_1,\dots,a_k\}$. The $\mathbf{CRB}(\a)$ matrix would then be obtained by inverting this matrix. Below we use the explicit dependence of the estimator $\bm\Theta(\a)$ on its parameter vector $\a$ to calculate the $\mathbf{CRB}(\bm\Theta)$. We then bound the Mean Squared Error $\mathbf{MSE}(\bm\Theta)$ of any unbiased estimator by the trace of the $\mathbf{CRB}(\bm\Theta)$ matrix.

As we have already mentioned above the negative log-likelihood (\ref{neg_ll}) of the CAE population is not sensitive to the scaling of the shape matrix. Thus, the $\mathbf{FIM}$ is singular and cannot be inverted to obtain the $\mathbf{CRB}$, this phenomenon is known as non-identifiability of parameters, see e.g. \cite{li2012interpretation} and references therein for an extensive treatment of this issue. Indeed, it is impossible to estimate the scaling of the covariance due to the normalization in our model. Instead, we need an alternative parameterization which eliminates this scale invariance. Specifically, we add the constraint $\Tr{\bm\Theta} = p$ and reparameterize the structure of $\mathcal{L}$ while lowering its dimension to $k'=k-1$:
\begin{equation}
\mathcal{L}' = \D_0 + \left\{ \sum_{i=1}^{k-1} a_i \D_i,\D_i \in \mathbb{C}^{p \times p}, a_i  \in \mathbb{R}\right\}, \D_0 \succeq 0.
\label{affine_param_1}
\end{equation}
From now on we denote
\begin{equation*}
\mathcal{S}' = \mathcal{L}' \cap \mathcal{P}(p).
\end{equation*}
For example, in the Toeplitz and banded examples discussed above we have:
\begin{itemize}
\item {\bf{Toeplitz}}: The coefficient $a_1$ in $\mathcal{L}$ is no longer needed since the main diagonal becomes known, and
\begin{align*}
\D_0 = \I, \D_i = \B_{i+1}, i=1,\dots,2p-2.
\end{align*}
\item {\bf{Banded}}: Of the $p$ diagonal elements, the first $p-1$ are chosen as independent, and we obtain
\begin{align*}
&\D_0 = \I, \D_i = \B_i-\B_p, i=1,\dots,p-1, \\
&\D_{m} = \B_{m+1}, m=p,\dots,p(2b+1)-b(b+1)-1.
\end{align*}
\item {\bf{Direction of Arrival Problem}}: In the DOA case it is more convenient to set $\Tr{\bm\Theta} = \sigma^2p$, thus
\begin{equation*}
\D_0=\sigma^2\I.
\end{equation*}
All the matrices $\B_i, i=1,\dots,N$ satisfy $\Tr{\B_i} =  \b(\overline\theta_i)^H\b(\overline\theta_i) = p$, thus we set
\begin{equation*}
\D_i=\B_i-\I, i=1,\dots,N.
\end{equation*}
Note that $\D_i$ are linearly independent due to the specific choice of $\B_i$ as above.
\end{itemize}
Given this scale dependent parametrization, the $\mathbf{FIM}(\a)$ computed element-wise reads as
\begin{align}
&\mathbf{FIM}_{hm}(\a) = \mathbb{E}\(\frac{\partial \ln p(\x; \a_0)}{\partial a_h}\frac{\partial \ln p(\x; \a_0)}{\partial a_m}\) \nonumber \\
&= -\mathbb{E}\(\frac{\partial^2 \ln p(\x; \a_0)}{\partial a_h\partial a_m}\),
\label{def_hes_fim}
\end{align}
where $\a_0$ corresponds to the parametrization of the true covariance matrix.
We follow the calculations of \cite{besson2013fisher, greco2013cramer} to obtain
\small
\begin{align*}
&\mathbf{FIM}(\a)_{hm} =\frac{p\Tr{\bm\Theta_0^{-1}\D_m\bm\Theta_0^{-1}\D_h}-\Tr{\bm\Theta_0^{-1}\D_h}\Tr{\bm\Theta^{-1}\D_m}}{p+1}, \\
&h,m=1,\dots,k'.
\end{align*}
\normalsize
The unknown covariance $\bm\Theta$ depends on $\a$ linearly:
\begin{equation*}
 \vec{\bm\Theta} = \J \a,
\end{equation*}
where
\begin{equation*}
 \J = \left\{\frac{\partial \vec{\bm\Theta}}{\partial \a}\right\} = \{\vec{\D_1},\dots,\vec{\D_k}\}
\end{equation*}
is the Jacobian $p^2 \times k'$ matrix. Thus, the $\mathbf{CRB}$ of the covariance error reads as
\begin{equation*}
\mathbf{CRB}(\bm\Theta) = \J\:\mathbf{CRB}(\a)\J^H= \J\:\mathbf{FIM}^{-1}(\a)\J^H,
\end{equation*}
and the total $\mathbf{MSE}$ over all the elements in the covariance estimator is bounded as
\begin{equation*}
\mathbf{MSE}(\bm\Theta^{\text{Est}}) = \mathbb{E}\[\norm{\bm\Theta^{\text{Est}} - \bm\Theta_0}_F^2\] \geq \Tr{\mathbf{CRB}(\bm\Theta)}.
\end{equation*}

\section{Existing solutions}
In this section, we review the existing solutions to the covariance estimation problem with and without structure.

\subsection{Sample Covariance}
The classical solution to the above covariance estimation problem is the sample covariance matrix defined by
\begin{equation}
\bm\Theta^{\text{SC}} = \frac{1}{n} \sum_{i=1}^n \x_i \x_i^H.
\label{c_sam}
\end{equation}
The sample covariance estimator is unbiased, always exists and is asymptotically consistent in any distribution with bounded second moments by the Law of Large Numbers. In the Gaussian case when $n\geq p$, it also maximizes the likelihood and is asymptotically efficient. In the elliptical case it converges to a scaled shape matrix. Sample covariance has been extensively studied so far and is generally suboptimal. A broad exposition on sample covariance performance for a large class of distributions was performed in \cite{vershynin2012close, srivastava2011covariance}. An additional disadvantage of this estimator is its ignorance to the prior structure.

In the recent years there have been proposed a number of covariance matrix estimators for Gaussian models with convex structure based on sample covariance, see e.g. \cite{rothman2008sparse, shah2012group}.

\subsection{Tyler's M-estimator}
The most popular approach to the shape matrix estimation in elliptical distributions is due to Tyler \cite{tyler1987distribution}. Tyler's M-estimator estimator is defined as the fixed point solution to the equation:
\begin{equation}
\bm\Theta^{\text{Tyler}} = \frac{p}{n} \sum_{i=1}^n \frac{\x_i \x_i^H}{\x_i^H {\left[\bm\Theta^{\text{Tyler}}\right]}^{-1} \x_i}.
\label{tyler_formula}
\end{equation}
This equation defines $\bm\Theta^{\text{Tyler}}$ up to a scaling factor, so the scale has to be fixed by some additional constraint. Two popular choices are to fix $\Tr{\bm\Theta^{\text{Tyler}}}$ or $\det{\bm\Theta^{\text{Tyler}}}$ as constant. When $n>p$, it has been proven that the fixed point iteration converges to the unique solution with probability one \cite{maronna1976robust}. This estimator is asymptotically consistent in all elliptical distributions. In fact, it maximizes the likelihood of the CAE population (\ref{neg_ll}). The advantages of Tyler's estimator are its simplicity and robustness. Its most significant drawbacks are that it does not necessary exist if $n<p$ and can hardly exploit known structure since the optimization problem obtained by minimizing the empirical likelihood (\ref{neg_ll}) is not convex. In \cite{bandiera2010knowledge} knowledge based variants of the fixed point iteration were proposed without convergence analysis. Recently, regularized and structured versions of Tyler's estimator were proposed in \cite{abramovich2007time,chen2011robust,wiesel2012unified,wiesel2012geodesic} based on the theories of concave Perron Frobenius and geodesic convexity.
Another approach of imposing linear symmetry structure on Tyler's estimator, making extensive use of the $g$-convexity of the problem, was proposed by \cite{soloveychik2013group}, where the constraint set is given as a set of fixed points of certain isometries over the manifold $\mathcal{P}(p)$.
Unfortunately, these approaches are limited in their modeling capabilities and cannot deal with general convex models as described above.

\subsection{Convex Projection}
A natural approach for introducing convex structure into covariance estimation is via projection. In our settings the projection is made onto a convex set $\mathcal{S}'$ defined above, e.g. \cite{ottersten1998covariance,henrion2012projection}. Given any unstructured estimator $\bm\Theta^{\text{Est}}$, e.g., the sample covariance or Tyler's estimator, its projection onto the closed convex set $\mathcal{S}'$ is defined as
\begin{equation}
\mathcal{P}_{\mathcal{S}'}(\bm\Theta^{\text{Est}})= \argmin_{\M \in \mathcal{S}'}\norm{\M-\bm\Theta^{\text{Est}}}, \label{proj_est}
\end{equation}
where $\norm{\cdot}$ is some norm. For convex structures as described above, the projection is a convex optimization problem which can be efficiently solved using standard numerical packages, e.g., CVX, \cite{cvx, gb08}.

The main advantage of the projection method is that, when $\bm\Theta_0 \in \mathcal{S}'$, the projection $\mathcal{P}_{\mathcal{S}}(\bm\Theta^{\text{Est}})$ is closer to $\bm\Theta_0$ than $\bm\Theta^{\text{Est}}$. The main disadvantage is that it requires a two-step solution which does not couple the distribution properties and the structure information simultaneously and is therefore suboptimal.

\section{COCA Estimator}
\subsection{Definition}
In this section we propose COCA - the COnvexly ConstrAined covariance estimator for GE distributions. Unlike the existing solutions, COCA exploits both the elliptical nature and the structure of the underlying distribution. COCA is based on the GMM \cite{matyas1999generalized} together with an asymptotically tight convex relaxation.

 The underlying principle behind COCA is the following identity  \cite{frahm2004generalized, frahm2010generalization}:
\begin{equation}
\E \left(p \frac{\x_i \x_i^H}{\x_i^H \bm\Theta_0^{-1} \x_i}\right) = \bm\Theta_0,
\label{main_lem}
\end{equation}
holding for all GE and, in particular, CAE populations. Indeed, Tyler's estimator is just the sample based solution that satisfies this identity. When the number of samples is small, even without any structural assumptions, the solution to this equation does not necessarily exist. Instead, we propose the GMM approach which seeks an approximate solution to
\begin{equation}
\underset{\bm\Theta\in\mathcal{S}'}{\text{min}}
\norm{\bm\Theta-\frac{p}{n}\sum_{i=1}^n \frac{\x_i \x_i^H}{\x_i^H \bm\Theta^{-1} \x_i}},
\label{gmm_pr}
\end{equation}
where $\norm{\cdot}$ is some norm. Intuitively, this optimization tries to simultaneously solve Tyler's program and project it onto the set of prior structure. By choosing an adaptive weighted norm, an optimal solution to (\ref{gmm_pr}) would result in an asymptotically consistent and accurate estimator \cite{matyas1999generalized,ottersten1998covariance}. Unfortunately, the objective is non-convex and it is not clear how to find its global solution in a tractable manner.

In what follows, we propose a convex relaxation of (\ref{gmm_pr}) that allows a computationally efficient solution. First, let us introduce auxiliary variables $d_i,i=1,\dots,n$:
\begin{equation}
\begin{aligned}
& \underset{\bm\Theta\in\mathcal{S}',d_i}{\text{min}}
& & \norm{\bm\Theta-\frac{1}{n}\sum_{i=1}^n d_i \x_i \x_i^H} \\
& \text{subject to}
& & d_i = \frac{p}{\x_i^H \bm\Theta^{-1} \x_i}, i=1 \dots n.
\end{aligned}
\end{equation}
This problem is not convex due to the equality constraints. We suggest to relax them to the inequalities:
\begin{equation}
\begin{aligned}
& \underset{\bm\Theta\in\mathcal{S}',d_i}{\text{min}}
& & \norm{\bm\Theta-\frac{1}{n}\sum_{i=1}^n d_i \x_i \x_i^H} \\
& \text{subject to}
& & d_i \leq \frac{p}{\x_i^H \bm\Theta^{-1} \x_i}, i=1 \dots n, \\
& & & d_i > 0, i=1 \dots n.
\end{aligned}
\label{coca_program}
\end{equation}
This relaxed problem is actually a convex minimization program. In order to show this we use
\begin{prop}(Schur's Complement \cite{zhang2005schur})
For any hermitian matrix $\X$ of the form
\begin{equation*}
\X =
 \begin{pmatrix}
  \A & \B \\
  \B^H & \C
 \end{pmatrix},
\end{equation*}
if $\A$ and $\C$ are invertible then the following properties hold:
\begin{enumerate}
  \item $\X \succeq 0$ iff $\C \succ 0$ and $\A-\B\C^{-1}\B^H \succeq 0$,
  \item $\X \succeq 0$ iff $\A \succ 0$ and $\C-\B^H\A^{-1}\B \succeq 0$.
\end{enumerate}
\end{prop}
As a corollary we obtain that for  $\bm\Theta \in \mathcal{P}(p), \bm\Theta \succ 0, \x \in \mathbb{C}^p$ and $\alpha > 0$ the following conditions are equivalent:
\begin{enumerate}
  \item $\bm\Theta \succeq \frac{1}{\alpha} \x \x^H$,
  \item $\alpha \geq \x^H \bm\Theta^{-1} \x$.
\end{enumerate}
Thus, we rewrite the inequalities $d_i \leq \frac{p}{\x_i^H \bm\Theta^{-1} \x_i}, i=1 \dots n$ as linear matrix inequalities (LMI):
\begin{equation}
\bm\Theta^{\text{COCA}} = \arg \left\{
\begin{aligned}
& \underset{\bm\Theta\in\mathcal{S}',d_i}{\text{min}}
& & \norm{\bm\Theta-\frac{1}{n}\sum_{i=1}^n d_i \x_i \x_i^H} \\
& \text{subject to}
& & \bm\Theta \succeq \frac{1}{p} d_i \x_i \x_i^H, \forall i=1 \dots n,\\
& & & d_i > 0, \forall i=1 \dots n.
\end{aligned}
\right.
\label{sdp_f}
\end{equation}

In this form COCA can be efficiently computed by standard semi-definite program solvers, e.g., CVX, \cite{cvx, gb08}.

\subsection{Consistency}
The non-relaxed version of COCA in (\ref{gmm_pr}) is clearly a reasonable approach for structured covariance estimation in elliptical models. The interesting question is how tight is the relaxation. We now provide two promising results in this direction.

\begin{theorem} \label{theorem1}
In the unstructured case $\mathcal{S} = \mathcal{P}(p)$ with $n \geq p+1$, the COCA estimator is unique up to a positive scaling factor and coincides with Tyler's estimator.
\begin{proof}
It is known that when $n \geq p+1$, (\ref{sdp_f}) has at least one solution which results in a zero objective value. It is Tyler's estimator which satisfies
\begin{equation}
d^*_i = \frac{p}{\x_i^H \bm[\Theta^*]^{-1} \x_i}, i = 1 \dots n. \nonumber
\end{equation}
%These equalities hold up to a scaling factor.
%, which is chosen to satisfy the (\ref{main_prob_tr}) constraint.

It remains to show that there are no other feasible solutions which result in a zero objective and is not a scaled version of this one. Indeed, assume in contradiction that there is such an additional solution, and for it $\bm\Theta=\frac{1}{n} \sum_{i=1}^n d_i \x_i \x_i^H$. Multiply each inequality $d_i \leq \frac{p}{\x_i^H \bm\Theta^{-1} \x_i}$ by the matrix $\x_i \x_i^H$ for $i=1 \dots n$ and sum up to obtain
\begin{equation}
\bm\Theta = \frac{1}{n} \sum_{i=1}^n d_i \x_i \x_i^H \preceq \frac{p}{n} \sum_{i=1}^n \frac{\x_i \x_i^H}{\x_i^H \bm\Theta^{-1} \x_i} = f(\bm\Theta). \label{tyl_ineq}
\end{equation}
The inequality (\ref{tyl_ineq}) reads now as $\bm\Theta \preceq f(\bm\Theta)$.
As stated in the Corollary V.I from \cite{pascal2008covariance} (the Corollary V.I is formulated there for the real case, but it remains valid in the complex case as explained there by the authors), this implies that $\bm\Theta$ is the fixed point of $f$: $\bm\Theta = f(\bm\Theta)$, which is exactly the definition of Tyler's estimator in (\ref{tyler_formula}). Thus proving that it is the only solution to (\ref{sdp_f}) up to a positive scaling factor.
\end{proof}
\end{theorem}

Theorem 1 proves that the unconstrained COCA estimator performs as a classical MLE and is, thus, efficient with the asymptotically normal distribution:
\begin{equation*}
\sqrt{n}\left(\bm\Theta^{\text{COCA}} - \bm\Theta_0\right) \xrightarrow{P} \mathcal{N}(0, \mathbf{CRB}).
\end{equation*}
In the constrained case, for a general convex set $\mathcal{S}'$, the analysis is  more difficult but we still have promising asymptotic results.

\begin{theorem} \label{theorem2}
In the structured case, COCA is an asymptotically consistent estimator of the true shape matrix $\bm\Theta_0 \in \mathcal{S}'$.
\begin{proof}
Is provided in the Appendix.
\end{proof}
\end{theorem}
The efficiency of the constrained COCA estimator remains an open question. We do not expect the constrained estimator to be statistically efficient. However, we believe that it can get quite close by adaptively tuning the norm as detailed in the next subsection.

\subsection{Choice of norm}
\label{norm_sec}
The definition of the COCA estimator depends on the choice of the norm in the objective of (\ref{sdp_f}). The consistency result in Theorem \ref{theorem2} is invariant to this choice, but the finite sample performance may change significantly. Natural choices are the Frobenius, trace and spectral norms. In addition, it is well known from the theory of GMM that adaptive weighted norms can enhance the performance of estimators. Similar ideas were applied in \cite{ottersten1998covariance} to develop covariance matching estimators.

\section{Computational complexity}
The COCA estimator developed in this paper can significantly change the approach to Tyler's shape matrix estimator, as it transforms the involved optimization into a convex problem, making the imposition of affine constraints possible. Unfortunately, it suffers from a significant drawback since it is computationally complex and involves solution of high-dimensional non-linear programs. In fact, this is also the drawback in the Gaussian case, where the projection onto the structure set requires a comparable computational effort (see \cite{henrion2012projection} and references therein). The most appropriate general class of methods usually applied in such SDP programs is known as interior-point algorithms, \cite{vandenberghe1996semidefinite}, which are polynomial in the dimension of the problem.
In practice, interior-point optimizers like MOSEK, SeDuMi and SDPT3 solve problems in a fixed number of iterations between about $10$ and $100$. Each iteration has polynomial complexity typically $O((k+n)^3)$. The exact power and additional logarithmic multipliers depend on the norm involved in the COCA-optimization and other specific details of the program at hand. To enjoy the advantages of the COCA fully, the proposed algorithms should be tuned to exploit the specific structure of the problem, e.g. Toeplitz or banded and the norm. Because of this and lack of space such algorithms are outside the scope of the current paper. One of the main directions of our future research is the development of less demanding COCA solvers.

\section{Numerical results}

In this section we demonstrate the advantages of COCA using numerical simulations. We investigated the performance benefits of COCA when the true shape matrix was either Toeplitz, banded or constructed based on the DOA grid. We compared the following estimators: $\bm\Theta^{\text{SC}}$ in (\ref{c_sam}), $\bm\Theta^{\text{Tyler}}$ in (\ref{tyler_formula}), $\bm\Theta^{\text{Proj}}$ in  (\ref{proj_est}) and $\bm\Theta^{\text{COCA}}$ in (\ref{sdp_f}). In $\bm\Theta^{\text{Proj}}$ we projected Tyler's estimator when it existed and the sample covariance otherwise.

For each number of samples $n$ we generated $1000$ sets of independent, compound proper normally distributed $p$-dimensional samples and calculated the empirical $\mathbf{MSE}$ for all the estimators. The samples were generated as $\x = \sqrt{\tau} \mathbf{v}$, where the random variable $\tau \sim \chi^2$ and the random vector $\mathbf{v}$ was zero-mean circularly symmetric normally distributed with the covariance matrix $\bm\Theta_0$. To compare the performance of COCA and all the other estimators to the $\mathbf{CRB}$ we used the Frobenius norm in the optimization problem and when calculating the empirical $\mathbf{MSE}$.

\subsection{Toeplitz Covariance Matrix}
\label{toep}
\begin{figure}
\centering
\includegraphics[width=3.6in]{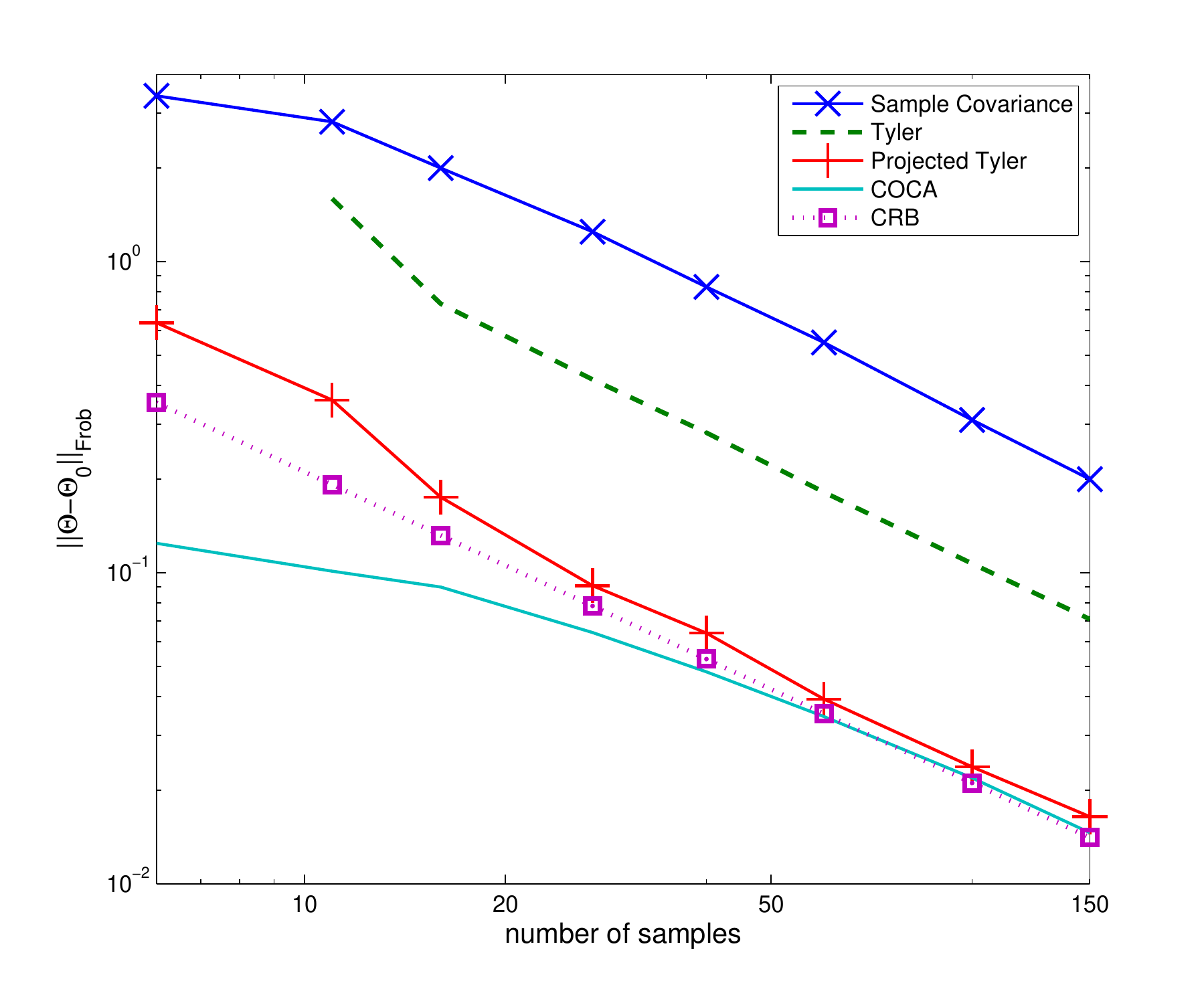}
\caption{COCA in the Toeplitz case.}
\label{pic_t}
\end{figure}
For $p=10$ the Toeplitz shape matrix was chosen to have $1$-s on the main diagonal and $\frac{1}{5}\pm\frac{j}{5}$, $\frac{1}{25}\pm\frac{j}{25}$ on the first two sub-diagonals correspondingly. The results are reported in Fig. \ref{pic_t}. It is easy to see the performance advantage of COCA over all the other estimators. For convenience we also put the constrained $\mathbf{CRB}$ for this case on the same plot. As the COCA estimator performs better in the sense of $\mathbf{MSE}$ we can imply that it is biased.

\subsection{Banded Covariance Matrix}
\begin{figure}
\centering
\includegraphics[width=3.6in]{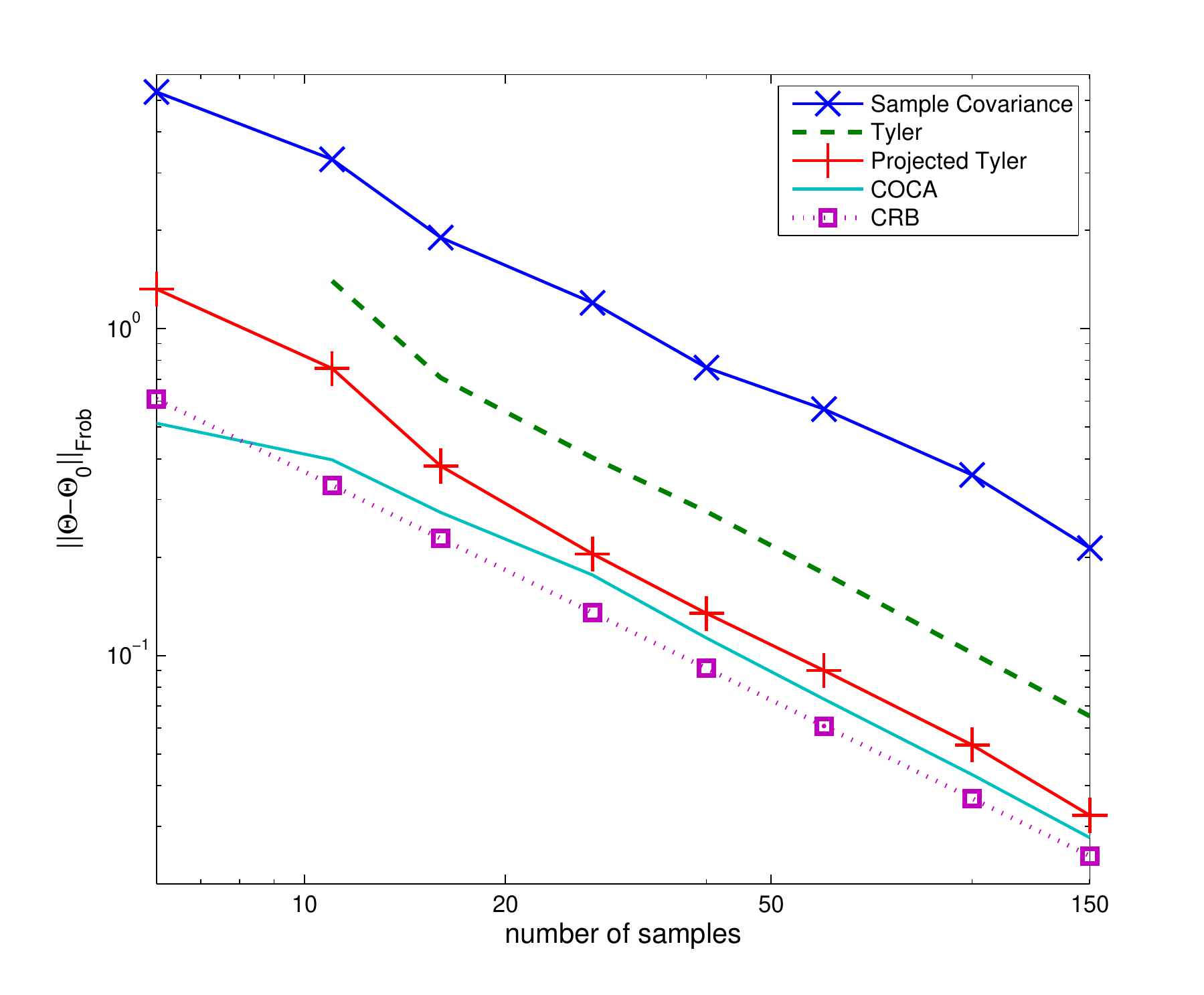}
\caption{COCA in the banded case.}
\label{pic_b}
\end{figure}
As an example of banded structure we took a matrix having the numbers $20,40,\dots,20p$ on the main diagonal for $p=10$, $12\pm 3j,\dots,(12\pm 3j)(p-1)$ and $2\pm 2j,\dots,(2\pm 2j)(p-2)$ on the first two sub-diagonals correspondingly and scaled it to have trace $p$. The band width is $2$ in this case. The averaged errors and the $\mathbf{CRB}$ are reported in Fig. \ref{pic_b}.

\subsection{DOA Covariance Matrix}
\begin{figure}
\centering
\includegraphics[width=3.6in]{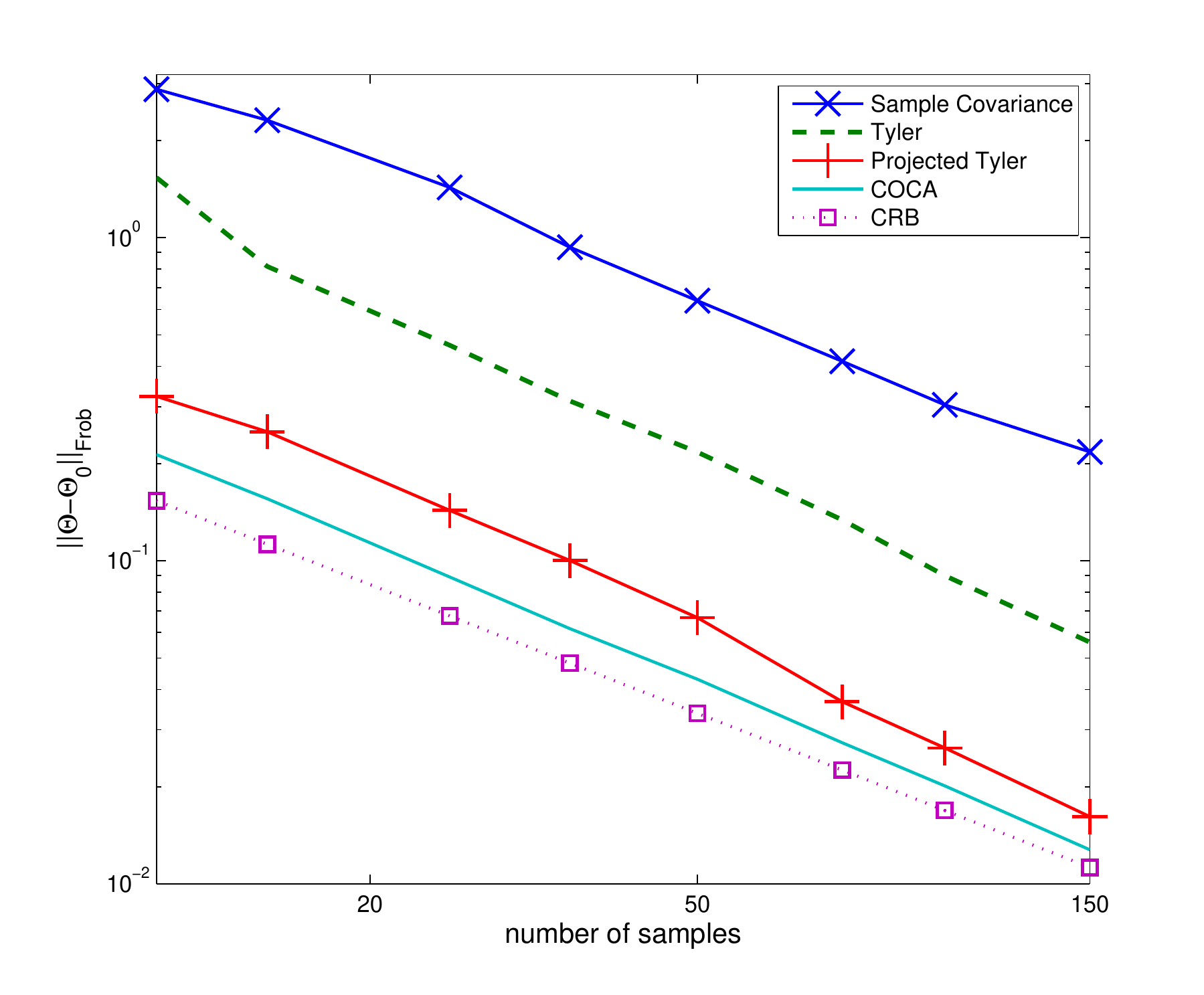}
\caption{COCA in the DOA case.}
\label{pic_d}
%\centering
%\includegraphics[width=3.3in]{huber.eps}
%\caption{Huber-COCA in the Toeplitz case.}
%\label{pic_h}
\end{figure}
In the DOA experiments we took $n = 5$ signal sources uniformly localized in the interval $[0,\pi]$. The noise added was white of energy $\sigma = 1/100$. The number of sensors was $p=10$. The grid was constructed by dividing the range $[0,\pi]$ into $p$ equal subintervals. The convergence rates for different estimators are present in Fig. \ref{pic_d}. We also provide the $\mathbf{CRB}$ for comparison.

\subsection{Different Norms Comparison}
\begin{figure}[!t]
\centering
\includegraphics[width=3.7in]{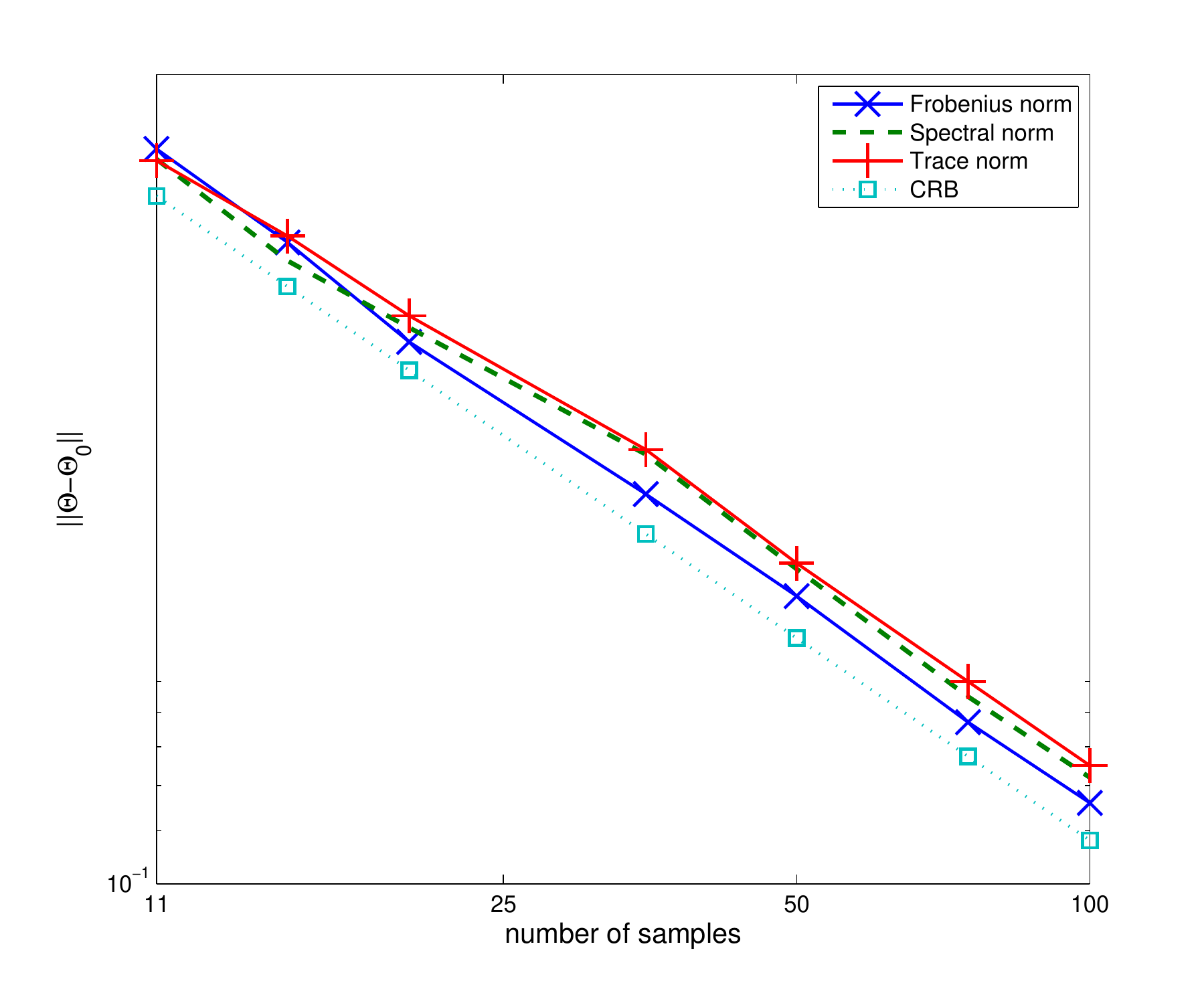}
\caption{Performance of the COCA with different norms in the banded case.}
\label{norm_comp}
\end{figure}

As we have already mentioned, the formulation of the COCA-estimator (\ref{sdp_f}) leaves freedom for the choice of the norm. In the examples above we used the Frobenius norm. In this section we compare the performance of COCA with different norms. In particular we took the spectral, the Frobenius and the trace norms and compared the COCA performance with the $\mathbf{CRB}$. The numerical results for the banded ($b=2$) type of constraints are provided in Figure \ref{norm_comp}. As we can see the choice of the norm affects the results quite slightly and the one making the optimization problem easier to solve should be picked.

\subsection{Discussion}
As we can see with the examples considered above, the COCA estimator outperformes the benchmarks used. Its $\mathbf{MSE}$ is actually quite close to the projection estimator, which can be considered as a one-step approximation to the COCA. We must also note that the $\mathbf{MSE}$ of COCA is less that the $\mathbf{CRB}$ for small values of $n$ in the first two figures. This is indeed possible, since COCA is only asymptotically unbiased and may turn out to be biased in finite samples.

\section{Conclusion}
In this paper we address structured covariance estimation in GE distributions. In particular, we assume that the covariance is a priori known to belong to a given convex set, e.g., the set of Toeplitz or banded matrices. We utilize the MLE of the shape matrix of normalized population which is a solution to a non-convex program and propose its convex relaxation based on the GMM technique. It is shown that the relaxed program (COCA) is tight in the unconstrained case and asymptotically tight in the constrained settings. Numerical simulations show that COCA performs better then other comparable techniques, such as unconstrained Tyler's estimator and its projection.

Our future work will first of all address the performance properties of the COCA estimator and its generalizations based on M-estimators. In addition, as we have already mentioned, when treated using general purpose numerical packages COCA may become a resource demanding program. Our second aim is to develop a more specific algorithm to make the COCA estimator computationally scalable.

\appendix

%\begin{rem}
%From the proof we see that the COCA-estimator makes the inequalities (\ref{main_prob_ineq}) tight, thus it always shows up to be the solution of (\ref{gmm_cov_gen}) and, thus of (\ref{tyler_formula}).
%\end{rem}

\subsection{Proof of Theorem \ref{theorem2}}
\begin{proof}
We assume that the set $\mathcal{S}' \subset \mathcal{P}(p)$ is a compact convex set separated from zero. Otherwise, we take a large enough ball $\mathcal{B}$ centered at zero and replace the constraint $\mathcal{S}'$ by the intersection $\mathcal{S}' \cap \mathcal{B}$. Denote $d_i^0=d_i^0(\x)=\frac{p}{\x_i^H \bm\Theta_0^{-1} \x_i}$. For the sake of convenience we will consider the realizations of the samples as infinite sequences $\x = (\x_1, \x_2, \dots, \x_n, \dots).$ Consider now a random function
\begin{equation*}
h_n (\bm\Theta_0, \d; \x) = \norm{\bm\Theta_0 - \frac{1}{n}\sum_{i=1}^n d_i \x_i \x_i^H},
\end{equation*}
where $\d = (d_1, d_2, \dots, d_n, \dots).$

The (strong) Law of Large Numbers implies that
\begin{align*}
&h_n (\bm\Theta_0, \d^0; \x) = \norm{\bm\Theta_0 - \frac{1}{n}\sum_{i=1}^n d_i^0 \x_i \x_i^H} \\
&\rightarrow \norm{\bm\Theta_0 - \E \left( \frac{\x_i \x_i^H}{\x_i \bm\Theta_0^{-1} \x_i} \right)}, n \rightarrow \infty \text{ } a.s.
\end{align*}
For the elliptical distribution $\E \left( \frac{ \x_i \x_i^H}{\x_i \bm\Theta_0^{-1} \x_i} \right) = \bm\Theta_0$ \cite{frahm2004generalized}, so we get that
\begin{equation}
h_n (\bm\Theta_0, \d^0; \x) \rightarrow 0, n \rightarrow \infty \text{ } a.s.
\label{true_conv}
\end{equation}
For now, given a realization $\x$ and a number $n \in \N$ denote by $(\widehat{\bm\Theta}^{(n)}, \widehat{\d}^{(n)})$ the solution of (\ref{coca_program}), all the $\widehat{d}^{(n)}_i = 0$ for $i > n$.
Define a random variable $\widetilde{h}_n$ depending on $\x$:
\begin{equation*}
\widetilde{h}_n (\x) = \norm{\widehat{\bm\Theta}^{(n)} - \frac{1}{n}\sum_{i=1}^n \widehat{d}^{(n)}_i \x_i \x_i^H}.
\end{equation*}
For each $\x$:
\begin{equation*}
\widetilde{h}_n (\x) \leq h_n (\bm\Theta_0, \d^0; \x),
\end{equation*}
since $(\widehat{\bm\Theta}^{(n)}, \widehat{\d}^{(n)})$ is the extremum of the target function, thus (\ref{true_conv}) implies
\begin{equation*}
\widetilde{h}_n (\x) \rightarrow 0, n \rightarrow 0 \text{ } a.s.
\end{equation*}
Since $\mathcal{S}'$ is compact we can choose a convergent subsequence, and renumber it if needed. We now have:
\begin{equation}
\widehat{\bm\Theta}^{(n)} \rightarrow \bar{\bm\Theta} \succ 0, n \rightarrow 0,
\label{c_conv}
\end{equation}
\begin{equation}
 \frac{1}{n}\sum_{i=1}^n \widehat{d}^{(n)}_i \x_i \x_i^H \rightarrow \bar{\bm\Theta}, n \rightarrow 0.
\label{d_conv}
\end{equation}
All these events happen with probability one and $\bar{\bm\Theta} = \bar{\bm\Theta}(\x)$ depends on the realization.
The relation (\ref{c_conv}) implies that  for any $\epsilon >0$ there exists $n_1 \in \N$ starting from which $\widehat{\bm\Theta}^{(n)} \prec (1 + \epsilon)\bar{\bm\Theta}$. Thus,
\begin{align}
&\frac{1}{n}\sum_{i=1}^n \widehat{d}^{(n)}_i \x_i \x_i^H \preceq \frac{1}{n}\sum_{i=1}^n \frac {\x_i \x_i^H}{\x_i^H \[\widehat{\bm\Theta}^{(n)}\]^{-1} \x_i} \nonumber\\
&\preceq \frac{1}{n}\sum_{i=1}^n \frac {\x_i \x_i^H}{\x_i^H {\bar{\bm\Theta}}^{-1} x_i} + \frac{\epsilon}{n}\sum_{i=1}^n \frac {\x_i \x_i^H}{\x_i^H {\bar{\bm\Theta}}^{-1} \x_i}.
\label{c_ineq}
\end{align}

In a similar way (\ref{d_conv}) implies that for the same $\epsilon >0$ there exists $n_2 \in \N$ starting from which
\begin{equation}
 (1-\epsilon)\bar{\bm\Theta} \preceq  \frac{1}{n}\sum_{i=1}^n \widehat{d}^{(n)}_i \x_i \x_i^H.
\label{m_conv}
\end{equation}
Now for $n \geq \max(n_1,n_2)$
\begin{align*}
&(1-\epsilon)\bar{\bm\Theta} \preceq  \frac{1}{n}\sum_{i=1}^n \widehat{d}^{(n)}_i \x_i \x_i^H \\
&\preceq \frac{1}{n}\sum_{i=1}^n \frac {\x_i \x_i^H}{\x_i^H {\bar{\bm\Theta}}^{-1} \x_i} + \frac{\epsilon}{n}\sum_{i=1}^n \frac {\x_i \x_i^H}{\x_i^H {\bar{\bm\Theta}}^{-1} \x_i}.
\end{align*}
Since $\epsilon$ was chosen arbitrarily and all the sums here are bounded, this implies that
\begin{equation}
 \bar{\bm\Theta} \preceq \frac{1}{n}\sum_{i=1}^n \frac {\x_i \x_i^H}{\x_i^H {\bar{\bm\Theta}}^{-1} \x_i},
\label{pascal_in}
\end{equation}
for sufficiently large $n$. From here the proof continues as in Theorem 1: due to Corollary V.I from \cite{pascal2008covariance} this implies that (\ref{pascal_in}) holds with equality and the uniqueness implies that $\bar{\bm\Theta}=\bm\Theta_0$ a.s. up to a scaling factor.

Assume now that the original sequence $\widehat{\bm\Theta}^{(n)}$ does not converge, which implies that it has a subsequence $\widehat{\bm\Theta}^{(n_i)}$, which converges to a different limit $\bar{\bm\Theta}_{\text{wrong}} \neq \bm\Theta_0$, but this contradicts the previous reasoning for the convergent subsequence, thus showing that the original sequence $\widehat{\bm\Theta}^{(n)}$ converges to $\bm\Theta_0$.
\end{proof}

\bibliographystyle{IEEEtran}
\bibliography{ilya_bib}

\begin{IEEEbiography}[{\includegraphics[width=1in,height=1.25in,clip,keepaspectratio]{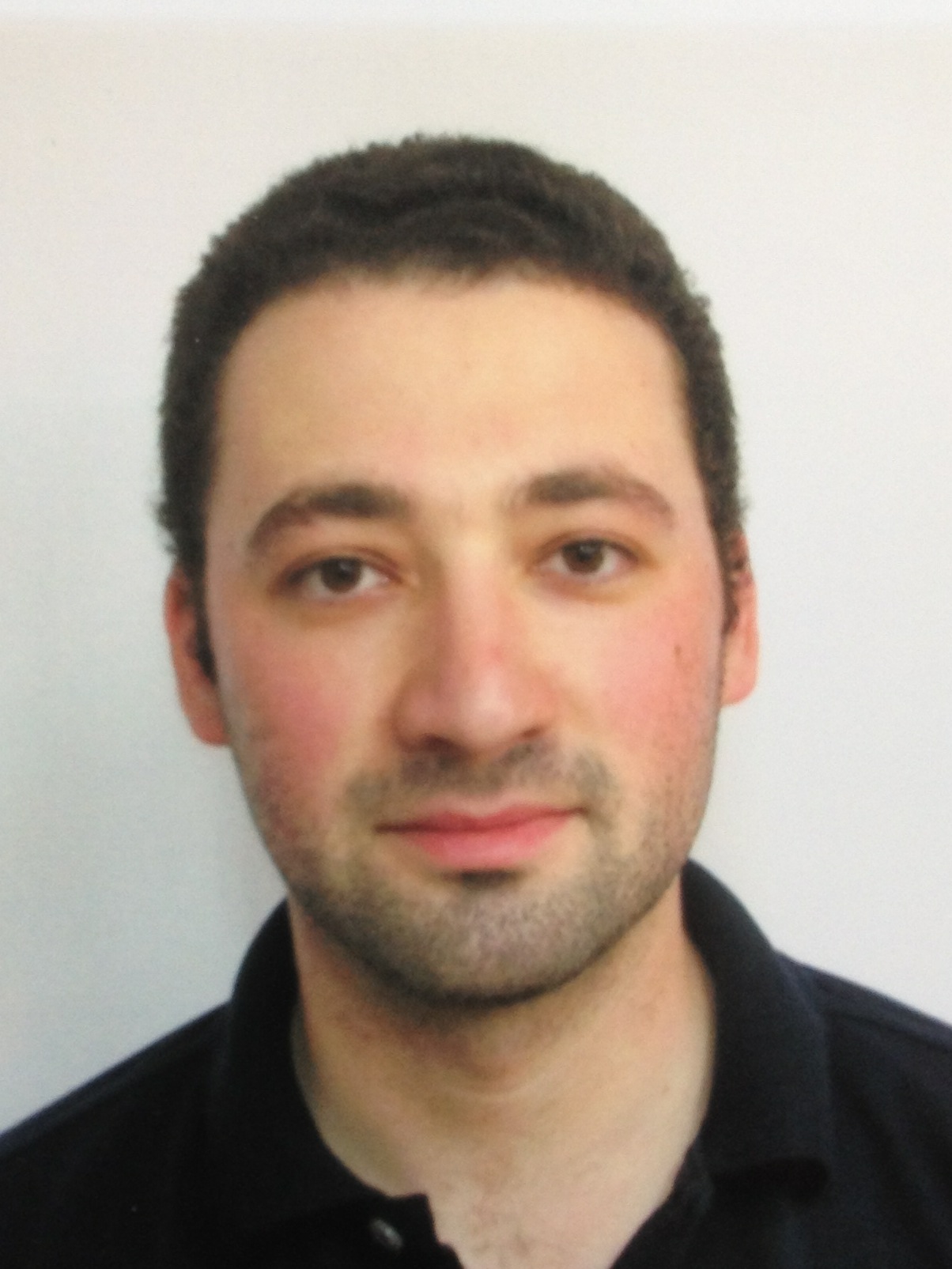}}]{Ilya Soloveychik}
(SM'13) received the B.Sc. degree in applied mathematics and physics from the Moscow Institute of Physics and Technology, in 2007, and the M.Sc degree in mathematics from the Hebrew University of Jerusalem, in 2013.
He is currently a Ph.D. student with the Rachel and Selim Benin School of Computer Science and Engineering, the Hebrew University of Jeursalem, Israel.

He was the Prize recipient of the Russian Olympiad in Physics in 2003. He received the Klein Prize and the Kaete Klausner Research Scholarship in 2011.
\end{IEEEbiography}

\begin{IEEEbiography}[{\includegraphics[width=1in,height=1.25in,clip,keepaspectratio]{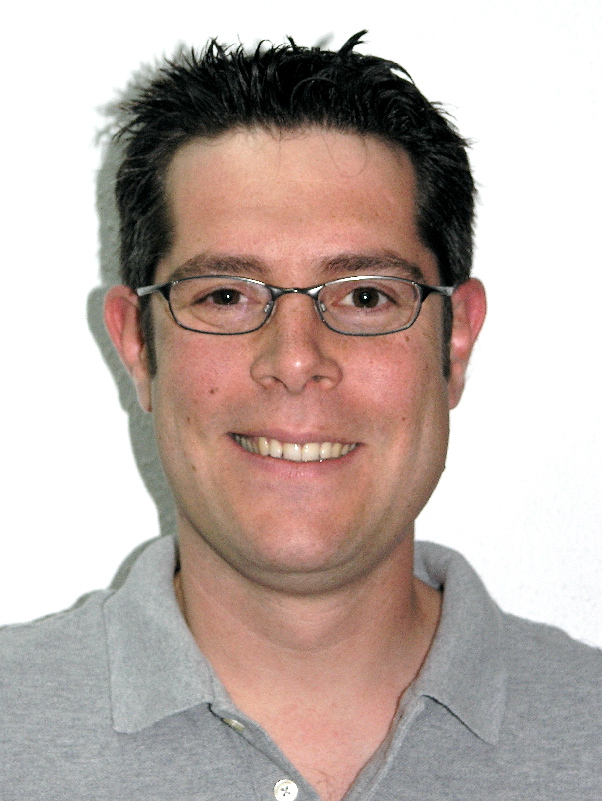}}]{Ami Wiesel}
received the B.Sc. and M.Sc. degrees in Electrical Engineering from Tel-Aviv University, Tel-Aviv, Israel, in 2000 and 2002, respectively, and the Ph.D. degree in Electrical Engineering from the Technion - Israel Institute of Technology, Haifa, Israel, in 2007. He was a postdoctoral fellow with the Department of Electrical Engineering and Computer Science, University of Michigan, Ann Arbor, in 2007-2009. Since Jan. 2010, he is a faculty member at the Rachel and Selim Benin School of Computer Science and Engineering at the Hebrew University of Jerusalem, Israel.

Dr. Wiesel was a recipient of the Young Author Best Paper Award for a 2006 paper in the IEEE Transactions on Signal Processing and a Student Paper Award for a 2005 Workshop on Signal Processing Advances in Wireless Communications (SPAWC) paper. He was awarded the Weinstein Study Prize in 2002, the Intel Award in 2005, the Viterbi Fellowship in 2005 and 2007, and the Marie Curie Fellowship in 2008.
\end{IEEEbiography}
\vfill
\end{document}